\documentclass[sigconf]{acmart}

\usepackage{booktabs} 
\usepackage{tikz}

\usepackage{algorithm}
\usepackage{algorithmic}
\floatstyle{plaintop}
\floatname{algorithm}{Algorithm}
\restylefloat{algorithm}
\usepackage{amsmath}
\usepackage{amssymb}

\setcopyright{rightsretained}
\renewcommand\footnotetextcopyrightpermission[1]{}
\acmDOI{10.475/123_4}

\acmISBN{123-4567-24-567/08/06}

\acmConference[GECCO '17]{the Genetic and Evolutionary Computation Conference 2017}{July 15--19, 2017}{Berlin, Germany}
\acmYear{2017}
\copyrightyear{2017}

\acmPrice{15.00}

\begin{document}
\title{Interval Arithmetic and Interval-Aware Operators for Genetic
  Programming}\titlenote{An abridged version of this paper appears in~\cite{dick2017intervals}}

\author{Grant Dick}
\affiliation{%
  \institution{Department of Information Science\\University of Otago}
  \city{Dunedin} 
  \state{New Zealand} 
}
\email{grant.dick@otago.ac.nz}

\renewcommand{\shortauthors}{G. Dick}

\begin{abstract}
  Symbolic regression via genetic programming is a flexible approach to
machine learning that does not require up-front specification of model
structure. However, traditional approaches to symbolic regression
require the use of protected operators, which can lead to perverse
model characteristics and poor generalisation. In this paper, we
revisit interval arithmetic as one possible solution to allow genetic
programming to perform regression using unprotected operators. Using
standard benchmarks, we show that using interval arithmetic within
model evaluation does not prevent invalid solutions from entering the
population, meaning that search performance remains compromised. We
extend the basic interval arithmetic concept with `safe' search
operators that integrate interval information into their process,
thereby greatly reducing the number of invalid solutions produced
during search. The resulting algorithms are able to more effectively
identify good models that generalise well to unseen data. We conclude
with an analysis of the sensitivity of interval arithmetic-based
operators with respect to the accuracy of the supplied input feature
intervals.

\end{abstract}

%
%



\maketitle

\setlength{\abovecaptionskip}{0pt}

\section{Introduction}
For over two decades, researchers have explored the use of genetic
programming (GP) to evolve models in a regression
setting~\cite{koza:book}. The resulting symbolic regression approach
is interesting as it simultaneously searches for a suitable model
structure and its corresponding parameters. This presents a highly
desirable framework that frees the user from a priori decisions
pertaining to model structure, allowing novel and potentially
insightful models to be discovered and provide better understanding of
the problem.

A long-known issue with symbolic regression is the need for protected
operators~\cite{koza:book}. Because we do not know up-front the
conditions under which a particular operation will be applied, we need
to ensure that its application will produce results that will not
corrupt any subsequent operations in the model. By far, the most
common solution to this is to develop customised `protected' operators
with built-in exception handling, such as for division by
zero. However, the very nature of GP is to exploit this behaviour to
create novel solutions that fit well to training data at the expense
of generalisation performance. An alternative solution that has been
proposed is to identify the ranges of input features and apply
interval arithmetic to identify safe and valid use of operators. These
intervals ideally capture knowledge of the problem domain that can
help identify whether or not a particular operation will remain valid
over all possible inputs. If interval arithmetic identifies an
operation as producing an undefined interval on some inputs, then we
can flag the individual containing that operation with a poor fitness
to discourage its selection and propagation.

Previous work has demonstrated some success in applying interval
arithmetic over protected operators~\cite{keijzer}. However, most work
incorporating interval arithmetic into GP uses it purely in an
evaluative framework --- the dynamics of interval arithmetic within
the population, and how it can be further exploited within GP, remain
largely unexplored. The goal of this paper is to explore how interval
arithmetic may be used to change the behaviour of GP by using interval
arithmetic to guide safe and effective use of search operators. In
doing so, it is shown that interval arithmetic, when used purely as an
evaluative tool, cannot prevent invalid individuals appearing in the
population during run. It is also shown that relying on selection
pressure to eliminate invalid solutions from the population is
ineffective, as standard mutation and crossover operators frequently
generate offspring that produce undefined execution intervals. New
search operators are presented in this paper that attempt to honour
the intervals presented by the problem domain to ensure that offspring
remain valid during search. This increases the rate of search within
the population, and leads to more rapid evolution towards fit
models. In addition to these new operators, the paper also examines
how effectively input intervals can be estimated from available data,
and to what extent GP using interval arithmetic is sensitive to the
quality of these estimated intervals.

The remainder of this paper is structured as follows:
\S\ref{sec:interval-arithmetic} provides an introduction to
interval arithmetic and related work in GP;
\S\ref{sec:basic-behaviour} demonstrates the dynamics of GP using interval arithmetic within evaluation, and informs the process
of developing new operators presented in \S\ref{sec:new-operators};
the new operators are examined using several benchmark
problems \S\ref{sec:benchmarks}, and the sensitivity of interval
arithmetic-based GP using estimated input intervals is also examined;
finally \S\ref{sec:conc} concludes the paper and suggests areas of
future work.

\section{Operator Safety in GP}\label{sec:interval-arithmetic}
Identification of models through symbolic regression is one of the
most thoroughly explored areas of GP research --- it is suggested that
over a third of GP research either uses symbolic regression problems
as benchmarks, or actively explores new ways in which to conduct
symbolic regression~\cite{bench}. Unlike traditional machine learning
approaches to regression, symbolic regression through GP does not
require the up-front selection of model structure. Instead, the GP
system is free to explore a range of possible model structures, giving
rise to the possibility of useful insights being discovered through
evolution. However, to achieve this flexibility, the GP must provide
closure, where the result of any operation must be valid as a possible
input to subsequent operations. This presents a problem to some
operators, such as division or computing the square root of a
number. When operators are used that do not define valid outputs over
all possible inputs, the GP system must incorporate some logic to
handle these exceptional cases.

\begin{table}
  \caption{Interval calculations for some of the operators used in
    this paper.}\label{tab:intervalops}
  \begin{tabular}{ll}
    \toprule
    {\bf Operator} & {\bf Resulting Interval} \\
    \midrule
    $[a, b]+[c, d]$ & $[a+c, b+d]$\\
    $[a, b]-[c, d]$ & $[a-d, b-c]$\\
    $[a, b]\times[c, d]$ & $[min(a {\times} b, a {\times} d, c {\times} b, c {\times} d), max(a {\times} b, a {\times} d, c {\times} b, c {\times} d)]$ \\
    $[a, b]\div[c, d]$ &  \begin{minipage}{0.1\linewidth}\[
      \begin{cases} 
        \mathrm{undefined} & \text{, if } 0 \in [c, d] \\
        \begin{split}
          [&min(a {\div} b, a {\div} d, c {\div} b, c {\div} d),\\
           &max(a {\div} b, a {\div} d, c {\div} b, c {\div} d)]\end{split} & \text{, otherwise}
      \end{cases}
      \]\end{minipage} \\
      $e^{[a, b]}$ & $[e^{a}, e^{b}]$ \\
      $log_{e}[a, b]$ & \begin{minipage}{0.1\linewidth}\[
        \begin{cases} 
          \mathrm{undefined} & \text{, if } a\leq 0 \\
                 [log_{e}a, log_{e}b] & \text{, otherwise}
        \end{cases}
        \]\end{minipage} \\
        \bottomrule
  \end{tabular}
\end{table}

\begin{figure}
  \centering
  \begin{minipage}{\linewidth}
    \begin{tikzpicture}[level/.style={sibling distance=0.667\linewidth/#1}]
      \node [circle,draw,label=0:{$[0.0,2.0]$}] (z){$\div$}
      child {node[draw=none,label=south:{$[0.0,1.0]$}] (l) {$x$}}
      child {node[circle,draw,label=0:{$[0.5,1.5]$}] (r) {$+$}
        child {node [draw=none,label=south:{$[0.5,0.5]$}] (rl) {$0.5$} }
        child {node [draw=none,label=south:{$[0.0,1.0]$}] (rr) {$y$} }
      }
      ;
    \end{tikzpicture}
  \end{minipage}
  \caption{Static analysis of a simple parse tree. Interval
    information percolates up the known intervals of the terminal
    nodes to the root of the tree.}
  \label{fig:interval}
\end{figure}
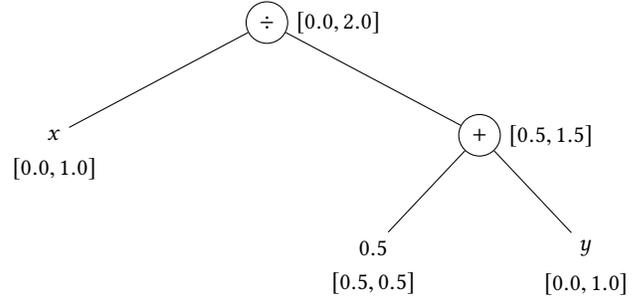

By far, the most common solution to the closure problem in symbolic
regression is to adopt protected versions of standard mathematical
operators. For example, the standard division operator can be replaced
by a protected version where any division by zero is replaced with the
constant value 1. This is the solution presented by Koza in his early
work exploring GP~\cite{koza:book}. While simple in nature and
implementation, the particular nuances of protected operators are
often exploited by GP to create solutions that fit well to training
data, but lead to poor generalisation. As an alternative to using
protected operators, Keijzer pioneered the idea of using information
about the known ranges of input variables to compute the expected
output interval of an evolved model~\cite{keijzer}. If we know the
interval of a variable (or equivalently a sub-expression), then we can
use simple rules to compute the expected interval of a mathematical
operation. Some examples of this are shown in
Table~\ref{tab:intervalops}. Following this, we can chain these
operations from the known intervals of our input features through to
the root of a given parse tree. An example of this is given in
Figure~\ref{fig:interval}: starting with the known intervals of our
terminals $x$ and $y$ (in this case, both defined over $[0,1]$), and
our constant 0.5, we can work out the interval of the addition
operator, and then finally the division at the root of the tree.

Since Keijzer's original paper on using interval arithmetic in GP,
interval arithmetic has seen use primarily as a tool to evaluate
individual solutions and assign fitness penalties to individuals that
present potentially invalid execution
intervals~\cite{Valigiani2004,Keijzer:2005:DEV:1068009.1068343,Tran:2016:DCM:2908961.2909002}. Kotanchek
et al. use interval arithmetic to help select intervals with robust
intervals as part of a multi-objective symbolic regression
framework~\cite{Kotanchek2008}. Others have extended the spirit of
interval arithmetic into using other forms of validation, such as
affine arithmetic~\cite{pennachin2010robust}. However, the influence
of interval arithmetic within the search process itself remains a
largely unexplored concept.

\section{Behaviour of Interval Arithmetic}\label{sec:basic-behaviour}
This paper aims to uncover some of the within-run properties of
interval arithmetic in GP with the aim of developing new operators
that exploit interval information to provide more effective search. We
start by exploring a simple  problem from previous
work~\cite{keijzer}. This function, referred to as {\em
  Keijzer-10} is defined over two variables:\begin{equation}
  f\left(x_{1}, x_{2}\right) = x_{1}^{x_{2}}
\end{equation}
where $x_{1}$ and $x_{2}$ are both defined over the interval
$[0,1]$. In previous work, this problem proved a challenge for GP,
where without interval arithmetic, GP failed to find a meaningful
solution in 98\% of runs. Here we will explore this problem again,
using a standard implementation of GP, once with protected operators
and then again with unprotected operators, and then an implementation
augmented with interval arithmetic to perform static evaluation. The
static evaluation was performed as follows: after creation through
initialisation, mutation or crossover, the known intervals of the
terminal nodes were supplied to the tree, and were recursively passed
through the tree to compute the solution's execution interval. If the
interval is deemed valid, then the individual is subsequently passed
on to a normal evaluation process. If the interval is deemed invalid,
then the individual is assigned the lowest possible fitness, with the
intended effect being to remove it from selection in the next
generation. The parameters for the GP system are defined in
Table~\ref{tab:parameters} and were settled upon after examining
recent work and some small experimentation trying different population
size and generations~\cite{dick2015improving,castelli2015c++}. As
interval arithmetic is attempting to characterise aspects of
generalisation performance, a hold-out set testing approach was used
similar to that in previous work. For each run, 20 instances were
sampled uniformly from the problem domain and used as a training
set. For testing, a mesh sample over the problem domain was used to
generate 10000 points uniformly over the problem space. In total, 100
separate runs were performed, each using a different training set. For
analysis, we explored three aspects, the training performance (in
terms of the best individual in the population at a given generation),
the performance of the best individual on the test set, and the number
of invalid individuals residing in the population in a given time
frame. Error performance was recorded using the root-relative squared
error (RRSE) measure:\begin{equation} RRSE(y,\hat{y})
= \sqrt{\frac{\sum_{i=1}^{|y|}\left(y_{i}
- \hat{y}_{i}\right)^{2}}{\sum_{i=1}^{|y|}\left(y_{i}
- \bar{y}\right)^{2}}},
\end{equation} where $y$ and $\hat{y}$ are the recorded response
values of the data set and the model predictions, respectively. The
RRSE is analogous to the normalised root mean square error used in
previous work. In all graphs presented in this paper, the statistic
being plotted is the median, with corresponding shaded areas
representing a 95\% confidence interval of the median.

\begin{table}
  \caption{Parameters used for all experiments in this
    paper.}\label{tab:parameters}
  \begin{tabular}{p{0.333\linewidth}p{0.667\linewidth}}
    \toprule
    {\bf Parameter} & {\bf Setting} \\
    \midrule
    Population size & 200 \\
    Generations & 250 \\
    Initialisation & Ramped Half-and-Half \\
    Min. initial depth & 2 \\
    Max. initial depth & 6 \\
    Mutation & Subtree (max. depth 4) \\
    Crossover & Subtree swap \\
    Max. offspring depth & 17 \\
    Crossover prob. & 0.3 \\
    Mutation prob. & 0.7 \\
    Selection & Tournament (size: 3) \\
    Elitism & 1 (fittest from previous generation)\\
    Function set & $+$, $-$, $\times$, $\div$, sin, cos, exp, log \\
    Terminal set & Input features: $x_{1}, x_{2}, \ldots, x_{p}$, plus
    ephemeral random constants from a uniform distribution over [-5, 5] \\
    \bottomrule
  \end{tabular}
\end{table}

\begin{figure}
  \centering
  \input{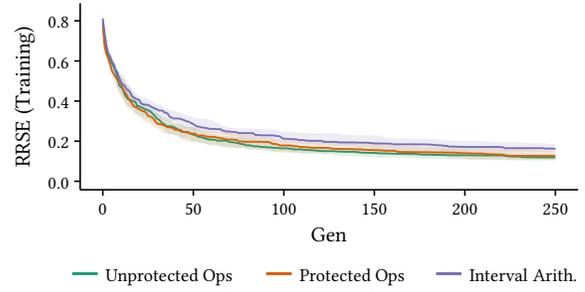}
  \caption{Training performance on the {\em Keijzer-10} problem.}
  \label{fig:keijzer10-train}
\end{figure}

\begin{figure}
  \centering
  \input{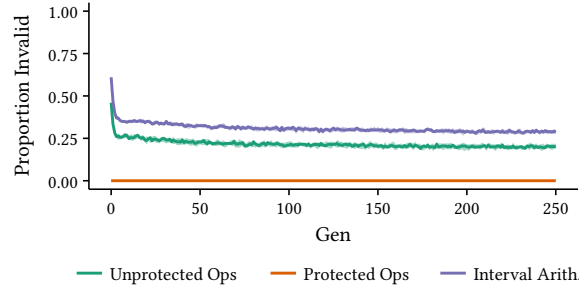}
  \caption{Proportion of population with invalid performance on the
    {\em Keijzer-10} problem.}
  \label{fig:keijzer10-inv}
\end{figure}

\begin{figure}
  \centering
  \input{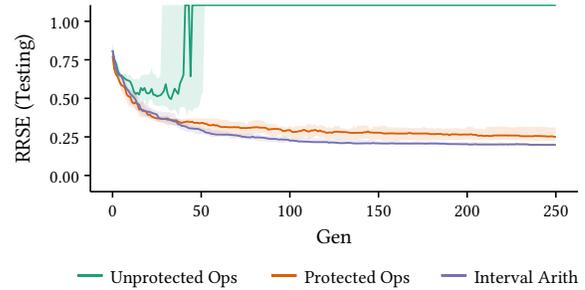}
  \caption{Test performance on the {\em Keijzer-10} problem.}
  \label{fig:keijzer10-test}
\end{figure}

The results of the initial analysis using {\em Keijzer-10} are shown
in Figures~\ref{fig:keijzer10-train}--\ref{fig:keijzer10-test}. The
results are rather interesting: in terms of training performance, both
protected and unprotected operators appear to allow the GP system to
evolve at a faster rate than when using interval arithmetic and static
analysis. A possible reason for this is provided by the results in
Figure~\ref{fig:keijzer10-inv}, which graphs the number of invalid
individuals present in the population in a given
generation.\footnote{We define an `invalid' individual here as either
  one that has produced errors while being evaluated on the training
  data, or one that was identified as potentially producing errors as
  a consequence of static analysis using interval arithmetic}. As can
be seen, the use of interval arithmetic means that more offspring are
identified as being invalid. Intuitively, this makes sense, as
interval arithmetic provides a second chance at flagging an individual
as problematic that may have been missed through the standard
evaluation process. However, the use of interval arithmetic has the
effect of reducing the size of the pool of individuals that selection
can effectively work on to produce the next generation, essentially
reducing the population size. It is interesting to note that the
proportion of invalid solutions in the population does not
substantially decrease over time.

\begin{figure}
  \centering
\begin{tikzpicture}[x=1pt,y=1pt]
\definecolor{fillColor}{RGB}{255,255,255}
\path[use as bounding box,fill=fillColor,fill opacity=0.00] (0,0) rectangle (216.81,108.41);
\begin{scope}
\path[clip] ( 32.37, 18.56) rectangle (216.81,108.41);
\definecolor{drawColor}{gray}{0.20}
\definecolor{fillColor}{gray}{0.20}

\path[draw=drawColor,line width= 0.4pt,line join=round,line cap=round,fill=fillColor] ( 66.95, 49.48) circle (  1.96);

\path[draw=drawColor,line width= 0.4pt,line join=round,line cap=round,fill=fillColor] ( 66.95, 50.73) circle (  1.96);

\path[draw=drawColor,line width= 0.6pt,line join=round] ( 66.95, 43.42) -- ( 66.95, 45.09);

\path[draw=drawColor,line width= 0.6pt,line join=round] ( 66.95, 39.54) -- ( 66.95, 34.63);
\definecolor{fillColor}{RGB}{27,158,119}

\path[draw=drawColor,line width= 0.6pt,line join=round,line cap=round,fill=fillColor] ( 45.34, 43.42) --
	( 45.34, 39.54) --
	( 88.57, 39.54) --
	( 88.57, 43.42) --
	( 45.34, 43.42) --
	cycle;

\path[draw=drawColor,line width= 1.1pt,line join=round] ( 45.34, 41.00) -- ( 88.57, 41.00);
\definecolor{fillColor}{gray}{0.20}

\path[draw=drawColor,line width= 0.4pt,line join=round,line cap=round,fill=fillColor] (124.59, 86.40) circle (  1.96);

\path[draw=drawColor,line width= 0.4pt,line join=round,line cap=round,fill=fillColor] (124.59, 73.92) circle (  1.96);

\path[draw=drawColor,line width= 0.4pt,line join=round,line cap=round,fill=fillColor] (124.59, 81.37) circle (  1.96);

\path[draw=drawColor,line width= 0.4pt,line join=round,line cap=round,fill=fillColor] (124.59, 81.60) circle (  1.96);

\path[draw=drawColor,line width= 0.4pt,line join=round,line cap=round,fill=fillColor] (124.59, 75.60) circle (  1.96);

\path[draw=drawColor,line width= 0.6pt,line join=round] (124.59, 53.90) -- (124.59, 69.03);

\path[draw=drawColor,line width= 0.6pt,line join=round] (124.59, 41.11) -- (124.59, 32.16);
\definecolor{fillColor}{RGB}{217,95,2}

\path[draw=drawColor,line width= 0.6pt,line join=round,line cap=round,fill=fillColor] (102.98, 53.90) --
	(102.98, 41.11) --
	(146.20, 41.11) --
	(146.20, 53.90) --
	(102.98, 53.90) --
	cycle;

\path[draw=drawColor,line width= 1.1pt,line join=round] (102.98, 44.87) -- (146.20, 44.87);
\definecolor{fillColor}{gray}{0.20}

\path[draw=drawColor,line width= 0.4pt,line join=round,line cap=round,fill=fillColor] (182.23, 55.79) circle (  1.96);

\path[draw=drawColor,line width= 0.4pt,line join=round,line cap=round,fill=fillColor] (182.23, 52.24) circle (  1.96);

\path[draw=drawColor,line width= 0.4pt,line join=round,line cap=round,fill=fillColor] (182.23, 52.59) circle (  1.96);

\path[draw=drawColor,line width= 0.4pt,line join=round,line cap=round,fill=fillColor] (182.23, 64.47) circle (  1.96);

\path[draw=drawColor,line width= 0.4pt,line join=round,line cap=round,fill=fillColor] (182.23, 51.47) circle (  1.96);

\path[draw=drawColor,line width= 0.4pt,line join=round,line cap=round,fill=fillColor] (182.23, 51.80) circle (  1.96);

\path[draw=drawColor,line width= 0.4pt,line join=round,line cap=round,fill=fillColor] (182.23, 52.88) circle (  1.96);

\path[draw=drawColor,line width= 0.4pt,line join=round,line cap=round,fill=fillColor] (182.23, 56.52) circle (  1.96);

\path[draw=drawColor,line width= 0.4pt,line join=round,line cap=round,fill=fillColor] (182.23, 54.55) circle (  1.96);

\path[draw=drawColor,line width= 0.4pt,line join=round,line cap=round,fill=fillColor] (182.23, 55.36) circle (  1.96);

\path[draw=drawColor,line width= 0.6pt,line join=round] (182.23, 43.87) -- (182.23, 48.27);

\path[draw=drawColor,line width= 0.6pt,line join=round] (182.23, 39.51) -- (182.23, 34.98);
\definecolor{fillColor}{RGB}{117,112,179}

\path[draw=drawColor,line width= 0.6pt,line join=round,line cap=round,fill=fillColor] (160.61, 43.87) --
	(160.61, 39.51) --
	(203.84, 39.51) --
	(203.84, 43.87) --
	(160.61, 43.87) --
	cycle;

\path[draw=drawColor,line width= 1.1pt,line join=round] (160.61, 41.11) -- (203.84, 41.11);
\definecolor{drawColor}{RGB}{0,0,0}

\node[text=drawColor,anchor=base,inner sep=0pt, outer sep=0pt, scale=  0.57] at ( 66.95, 20.69) {+ 82 (Max: Inf)};

\node[text=drawColor,anchor=base,inner sep=0pt, outer sep=0pt, scale=  0.57] at (124.59, 20.69) {+ 2 (Max: 1.347e+12)};
\end{scope}
\begin{scope}
\path[clip] (  0.00,  0.00) rectangle (216.81,108.41);
\definecolor{drawColor}{RGB}{0,0,0}

\path[draw=drawColor,line width= 0.6pt,line join=round] ( 32.37, 18.56) --
	( 32.37,108.41);
\end{scope}
\begin{scope}
\path[clip] (  0.00,  0.00) rectangle (216.81,108.41);
\definecolor{drawColor}{RGB}{0,0,0}

\node[text=drawColor,anchor=base east,inner sep=0pt, outer sep=0pt, scale=  0.77] at ( 28.19, 23.70) {0.00};

\node[text=drawColor,anchor=base east,inner sep=0pt, outer sep=0pt, scale=  0.77] at ( 28.19, 42.27) {0.25};

\node[text=drawColor,anchor=base east,inner sep=0pt, outer sep=0pt, scale=  0.77] at ( 28.19, 60.83) {0.50};

\node[text=drawColor,anchor=base east,inner sep=0pt, outer sep=0pt, scale=  0.77] at ( 28.19, 79.39) {0.75};

\node[text=drawColor,anchor=base east,inner sep=0pt, outer sep=0pt, scale=  0.77] at ( 28.19, 97.95) {1.00};
\end{scope}
\begin{scope}
\path[clip] (  0.00,  0.00) rectangle (216.81,108.41);
\definecolor{drawColor}{RGB}{0,0,0}

\path[draw=drawColor,line width= 0.6pt,line join=round] ( 30.12, 26.36) --
	( 32.37, 26.36);

\path[draw=drawColor,line width= 0.6pt,line join=round] ( 30.12, 44.92) --
	( 32.37, 44.92);

\path[draw=drawColor,line width= 0.6pt,line join=round] ( 30.12, 63.48) --
	( 32.37, 63.48);

\path[draw=drawColor,line width= 0.6pt,line join=round] ( 30.12, 82.05) --
	( 32.37, 82.05);

\path[draw=drawColor,line width= 0.6pt,line join=round] ( 30.12,100.61) --
	( 32.37,100.61);
\end{scope}
\begin{scope}
\path[clip] (  0.00,  0.00) rectangle (216.81,108.41);
\definecolor{drawColor}{RGB}{0,0,0}

\path[draw=drawColor,line width= 0.6pt,line join=round] ( 32.37, 18.56) --
	(216.81, 18.56);
\end{scope}
\begin{scope}
\path[clip] (  0.00,  0.00) rectangle (216.81,108.41);
\definecolor{drawColor}{RGB}{0,0,0}

\path[draw=drawColor,line width= 0.6pt,line join=round] ( 66.95, 16.31) --
	( 66.95, 18.56);

\path[draw=drawColor,line width= 0.6pt,line join=round] (124.59, 16.31) --
	(124.59, 18.56);

\path[draw=drawColor,line width= 0.6pt,line join=round] (182.23, 16.31) --
	(182.23, 18.56);
\end{scope}
\begin{scope}
\path[clip] (  0.00,  0.00) rectangle (216.81,108.41);
\definecolor{drawColor}{RGB}{0,0,0}

\node[text=drawColor,rotate= 90.00,anchor=base,inner sep=0pt, outer sep=0pt, scale=  0.90] at (  9.38, 63.48) {RRSE (Testing)};
\end{scope}
\begin{scope}
\path[clip] (  0.00,  0.00) rectangle (216.81,108.41);

\path[] ( 22.63, -0.00) rectangle (226.55,  4.93);
\end{scope}
\begin{scope}
\path[clip] (  0.00,  0.00) rectangle (216.81,108.41);
\definecolor{drawColor}{gray}{0.20}

\path[draw=drawColor,line width= 0.6pt,line join=round,line cap=round] ( 32.26,  1.20) --
	( 32.26,  3.01);

\path[draw=drawColor,line width= 0.6pt,line join=round,line cap=round] ( 32.26,  9.03) --
	( 32.26, 10.84);
\definecolor{fillColor}{RGB}{27,158,119}

\path[draw=drawColor,line width= 0.6pt,line join=round,line cap=round,fill=fillColor] ( 27.75,  3.01) rectangle ( 36.78,  9.03);

\path[draw=drawColor,line width= 0.6pt,line join=round,line cap=round] ( 27.75,  6.02) --
	( 36.78,  6.02);
\end{scope}
\begin{scope}
\path[clip] (  0.00,  0.00) rectangle (216.81,108.41);
\definecolor{drawColor}{gray}{0.20}

\path[draw=drawColor,line width= 0.6pt,line join=round,line cap=round] (105.60,  1.20) --
	(105.60,  3.01);

\path[draw=drawColor,line width= 0.6pt,line join=round,line cap=round] (105.60,  9.03) --
	(105.60, 10.84);
\definecolor{fillColor}{RGB}{217,95,2}

\path[draw=drawColor,line width= 0.6pt,line join=round,line cap=round,fill=fillColor] (101.09,  3.01) rectangle (110.12,  9.03);

\path[draw=drawColor,line width= 0.6pt,line join=round,line cap=round] (101.09,  6.02) --
	(110.12,  6.02);
\end{scope}
\begin{scope}
\path[clip] (  0.00,  0.00) rectangle (216.81,108.41);
\definecolor{drawColor}{gray}{0.20}

\path[draw=drawColor,line width= 0.6pt,line join=round,line cap=round] (169.84,  1.20) --
	(169.84,  3.01);

\path[draw=drawColor,line width= 0.6pt,line join=round,line cap=round] (169.84,  9.03) --
	(169.84, 10.84);
\definecolor{fillColor}{RGB}{117,112,179}

\path[draw=drawColor,line width= 0.6pt,line join=round,line cap=round,fill=fillColor] (165.33,  3.01) rectangle (174.36,  9.03);

\path[draw=drawColor,line width= 0.6pt,line join=round,line cap=round] (165.33,  6.02) --
	(174.36,  6.02);
\end{scope}
\begin{scope}
\path[clip] (  0.00,  0.00) rectangle (216.81,108.41);
\definecolor{drawColor}{RGB}{0,0,0}

\node[text=drawColor,anchor=base west,inner sep=0pt, outer sep=0pt, scale=  0.77] at ( 40.09,  3.37) {Unprotected Ops};
\end{scope}
\begin{scope}
\path[clip] (  0.00,  0.00) rectangle (216.81,108.41);
\definecolor{drawColor}{RGB}{0,0,0}

\node[text=drawColor,anchor=base west,inner sep=0pt, outer sep=0pt, scale=  0.77] at (113.43,  3.37) {Protected Ops};
\end{scope}
\begin{scope}
\path[clip] (  0.00,  0.00) rectangle (216.81,108.41);
\definecolor{drawColor}{RGB}{0,0,0}

\node[text=drawColor,anchor=base west,inner sep=0pt, outer sep=0pt, scale=  0.77] at (177.67,  3.37) {Interval Arith.};
\end{scope}
\end{tikzpicture}
  \caption{Test performance of solutions in the final generation on
    the {\em Keijzer-10} problem. The number under a given boxplot
    indicates the number of runs where an RRSE > 1 was encountered
    (and the number in brackets indicates the largest RRSE
    encountered).}
  \label{fig:keijzer10-final}
\end{figure}
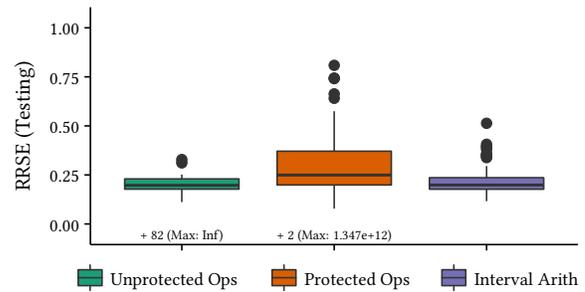

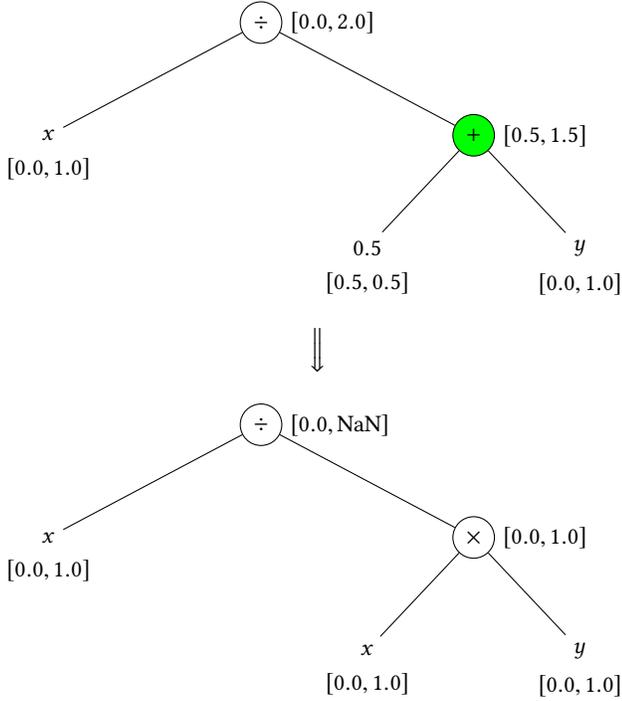
\begin{figure}
  \centering
  \begin{minipage}{\linewidth}
    \begin{tikzpicture}[level/.style={sibling distance=0.667\linewidth/#1}]
      \node [circle,draw,label=0:{$[0.0,2.0]$}] (z){$\div$}
      child {node[draw=none,label=south:{$[0.0,1.0]$}] (l) {$x$}}
      child {node[circle,draw,fill=green,label=0:{$[0.5,1.5]$}] (r) {$+$}
        child {node [draw=none,label=south:{$[0.5,0.5]$}] (rl) {$0.5$} }
        child {node [draw=none,label=south:{$[0.0,1.0]$}] (rr) {$y$} }
      }
      ;
    \end{tikzpicture}
  \end{minipage}
  
  \begin{minipage}{\linewidth}
    \vspace{1em}
    \begin{center}$\Big\Downarrow$\end{center}
    \vspace{1em}
  \end{minipage}
  
  \begin{minipage}{\linewidth}
    \begin{tikzpicture}[level/.style={sibling distance=0.667\linewidth/#1}]
      \node [circle,draw,label=0:{$[0.0,\mathrm{NaN}]$}] (z){$\div$}
      child {node[draw=none,label=south:{$[0.0,1.0]$}] (l) {$x$}}
      child {node[circle,draw,label=0:{$[0.0,1.0]$}] (r) {$\times$}
        child {node [draw=none,label=south:{$[0.0,1.0]$}] (rl) {$x$} }
        child {node [draw=none,label=south:{$[0.0,1.0]$}] (rr) {$y$} }
      }
      ;
    \end{tikzpicture}
  \end{minipage}
  \caption{Example of mutation producing an offspring with an invalid
    output interval.}
  \label{fig:mutation}
\end{figure}

While the integration of interval arithmetic into GP appears to
compromise training performance, it is clearly offset by stronger
generalisation performance. Examining Figures~\ref{fig:keijzer10-test}
and \ref{fig:keijzer10-final} show the good generalisation performance
provided by using interval arithmetic over protected and unprotected
operators. In agreement with previous work, unprotected operators
failed to find a meaningful solution to the problem in 82 out of 100
trials. Protected operators were also unable to properly manage this
problem. It is interesting to note, however, that on several
equations, GP using either protected or unprotected operators evolved a
solution that either matched, or very nearly matched, the following
equation:\begin{equation}
  f\left(x_{1}, x_{2}\right)=e^{\mathrm{log}_{e}\left(x_{1}\right) \cdot x_{2}}.
\end{equation} This particular equation is a perfect match for the
{\em Keijzer-10} problem on all inputs except for when $x_{1}=0$,
where it is undefined. Clearly, in these runs, there were no training
instances sampled with $x_{1}$ set to zero. Finding such a solution was
impossible for GP using interval arithmetic, regardless of training
set conditions, as such a solution would be ruled out during static
analysis.

\begin{algorithm}
  \caption{{\tt BuildTree}: the safe initialisation tree generation
    algorithm.}\label{alg:safe-init}
  \raggedright
  \begin{algorithmic}[1]
    \REQUIRE $depth$: the current tree depth; $maxdepth$: the maximum
    required tree depth; $F$: a list of functions for inner nodes;
    $T$: a list of terminals (the input features of the problem); $I$:
    the known intervals of the terminal choices (input features,
    constants, \ldots)
    \ENSURE A tree that produces output valid within the intervals
    defined by its subtrees
    
    \IF{{\tt PickTerminal}($depth$, $maxdepth$, $|F|$, $|T|$)}\label{alg:pickterm}
    \STATE $op \leftarrow$ random element from $T$
    \RETURN $\{root=op, left=\varnothing, right=\varnothing, interval=I[op]\}$
    \ELSE    
    \STATE $arg_{0} \leftarrow$ {\tt BuildTree}($depth+1$, $maxdepth$, $F$, $T$, $I$)\label{alg:buildsubtree}
    \STATE $ab \leftarrow arg_{0}[interval]$
    \IF{{\tt PickBinaryOperator}($F$)}
    \STATE $arg_{1} \leftarrow$ {\tt BuildTree}($depth+1$, $maxdepth$, $F$, $T$, $I$)
    \STATE $cd \leftarrow arg_{1}[interval]$
    \ELSE
    \STATE $arg_{1} \leftarrow \varnothing$
    \STATE $cd \leftarrow [\mathrm{NaN},\mathrm{NaN}]$
    \ENDIF
    \STATE $op \leftarrow$ {\tt SelectOperation}($F$, $ab$, $cd$)\label{alg:selectoperation}
    \STATE $I_{op} \leftarrow$ {\tt ComputeInterval}($op$, $ab$, $cd$)\label{alg:updateinterval}
    \RETURN $\{root=op, left=arg_{0}, right=arg_{1}, interval=I_{op}\}$
    \ENDIF
  \end{algorithmic}
\end{algorithm}

\begin{algorithm}
  \caption{{\tt CheckAndRepair}: walks up the parent nodes of the tree
    to ensure that each point in the tree produces a valid interval.}\label{alg:check-repair}
  \raggedright
  \begin{algorithmic}[1]
    \REQUIRE $node$ the parent node of the node that has just
    undergone a change (e.g., mutation or crossover); $F$: a list of
    functions for inner nodes;
    \ENSURE A tree that produces output valid within the intervals
    defined by its subtrees

    \STATE $ab \leftarrow node[left][interval]$
    \IF{{\tt BinaryOperator}($node[op]$)}
    \STATE $cd \leftarrow node[right][interval]$
    \ELSE
    \STATE $cd \leftarrow [\mathrm{NaN},\mathrm{NaN}]$
    \ENDIF
    \STATE $I_{op} \leftarrow$ {\tt ComputeInterval}($node[op]$, $ab$, $cd$)
    \IF{{\tt InvalidInterval}($I_{op}$)}
    \STATE shuffle $F$
    \FOR{$f \in F$}
    \STATE $I_{f} \leftarrow$ {\tt ComputeInterval}($f$, $ab$, $cd$)
    \IF{{\tt ValidInterval}($I_{f}$)}
    \STATE $node[op] \leftarrow f$
    \STATE $I_{op} \leftarrow I_{f}$
    \STATE {\bf break}
    \ENDIF
    \ENDFOR
    \ENDIF
    \STATE $node[interval] \leftarrow I_{op}$
    \STATE {\tt CheckAndRepair}($node[parent]$, $F$)
  \end{algorithmic}
\end{algorithm}

\begin{figure*}
  \centering
  \input{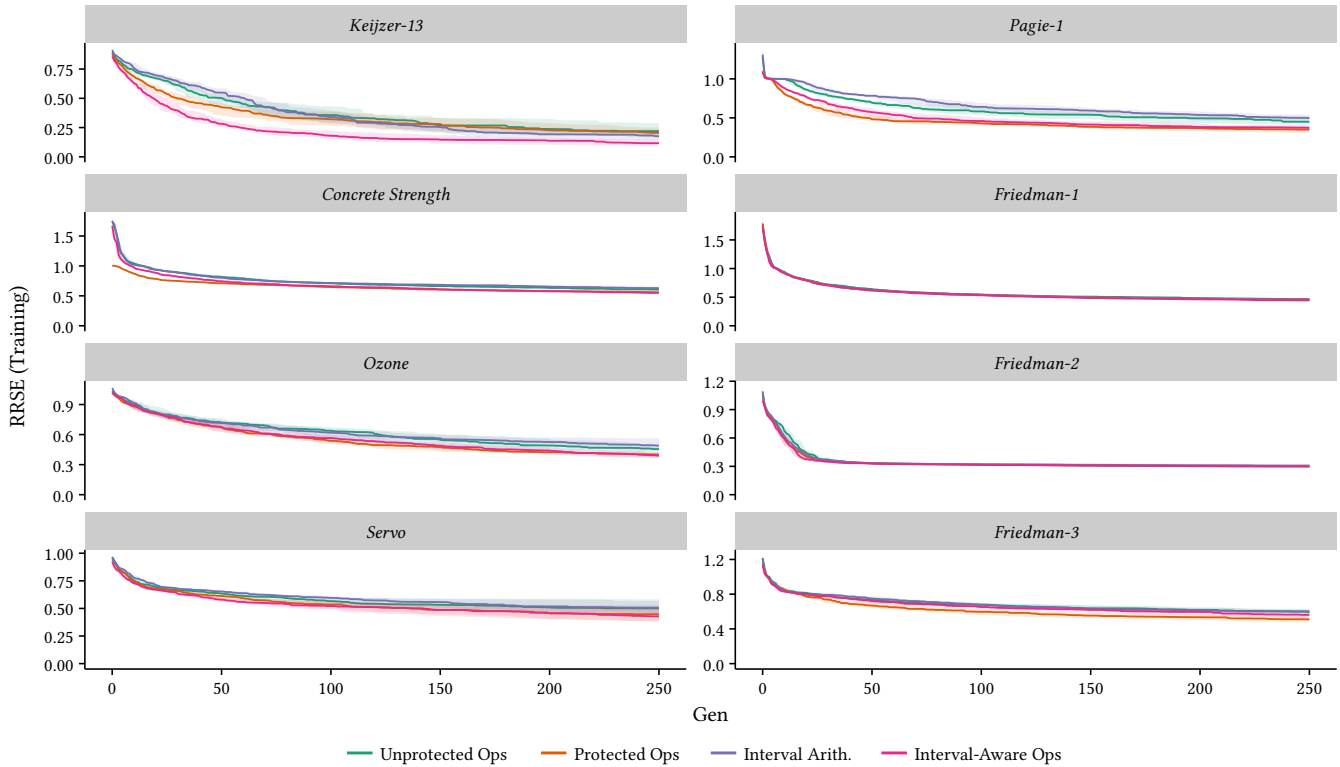}
  \caption{Training performance on selected benchmark problems.}
  \label{fig:bench-train}
\end{figure*}

\section{Interval-Preserving Operators}\label{sec:new-operators}

\begin{figure*}
  \centering
  \input{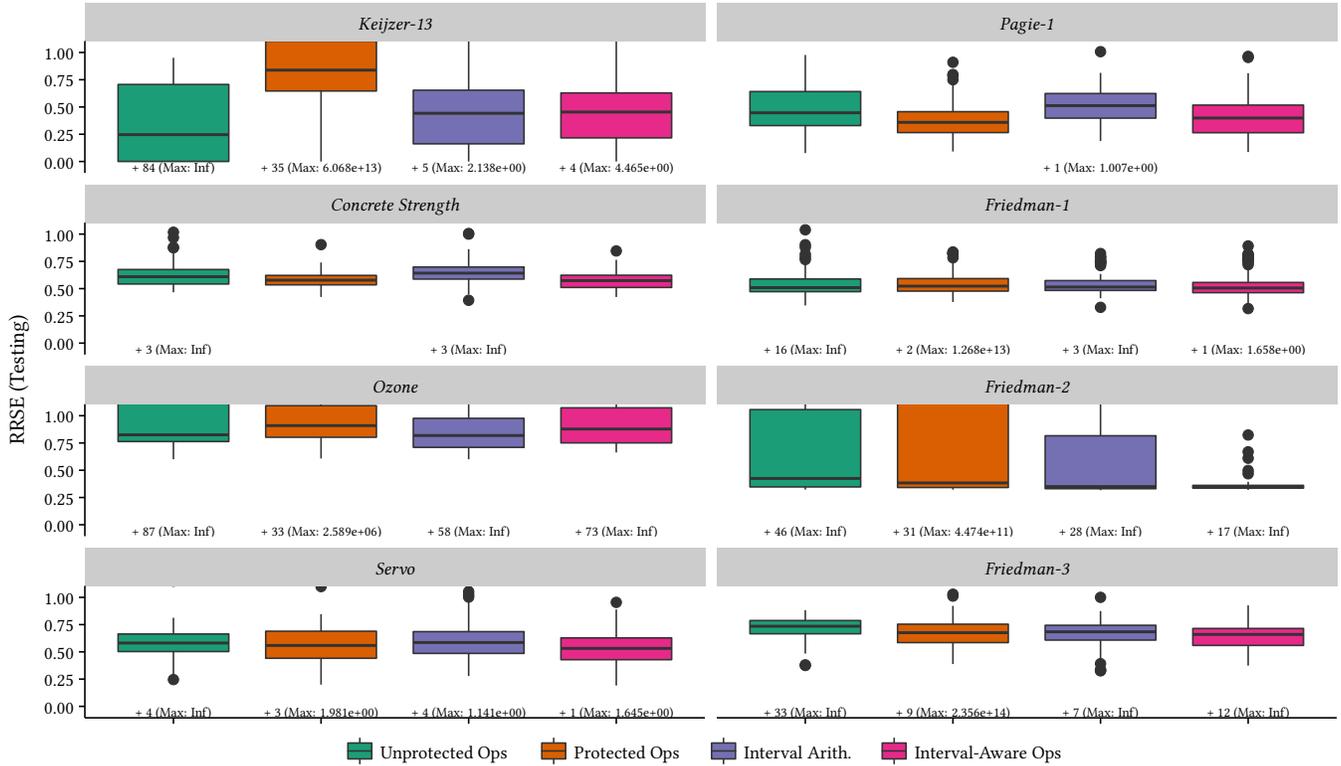}
  \caption{Test performance on selected benchmark problems.}
  \label{fig:bench-test}
\end{figure*}

The results presented in the previous section align with previous work
in suggesting that including interval arithmetic in GP can greatly
improve its generalisation performance. However, the training
performance results suggest that adopting interval arithmetic
invalidates more individuals during a run, which may reduce the
effective size of the population and thus require larger populations
and more generations to scale to harder problems. This remains an
issue as long as interval arithmetic is solely used for static
analysis within the the assignment of fitness to individuals. If the
information provided by interval arithmetic could also be used within
the search operators of GP, then this may result in fewer invalid
individuals, and so we may recover some of the lost search
performance.

Previous work identified that interval arithmetic could be used to
preserve working bounds of solutions developed through geometric
semantic genetic programming~\cite{dick2015improving}. The so-called
`safe initialisation' method modified the tree initialisation process
of GP such that the actual choice of operator at a given node was
delayed until all the necessary child nodes had been created. Once
subtrees were developed, their intervals could be computed and then
safe initialisation could select an appropriate operator to maintain a
valid interval for the entire tree. The end result is that invalid
solutions are prevented from entering the population. When compared to
an variant of GSGP that used logistic wrappers to manage intervals,
safe-initialisation GSGP evolved at a much faster
rate. Given that tree initialisation in GSGP is not
substantially different from standard GP, it should be reasonably
straightforward to integrate safe initialisation into standard GP. A
definition of safe initialisation is given in
Algorithm~\ref{alg:safe-init}: this variant differs slightly from
previous work by permitting both unary and binary operators
(the previous definition allowed only binary operators).

Ensuring valid solutions during tree initialisation is only one
aspect of operator protection that must be considered. In the previous
section, it was shown that the proportion of invalid solutions within
the population did not substantially change over time. This suggests
that the search operators themselves may be contributing to the
generation of invalid solutions. An example of how this may be
happening is shown in Figure~\ref{fig:mutation}. Here mutation is
applied to an individual with sane execution intervals. Likewise, the
interval of the new mutant subtree is well defined. However, when this
new tree is embedded into the solution, it interacts with its parent
node to produce an undefined execution interval. A similar scenario
can be imagined for crossover. What is needed, therefore, is a means
by which the operation above the swapped subtree can be considered
and, if necessary, repaired to produced valid intervals.

This paper extends the work done exploring safe initialisation --- in
addition to adopting safe initialisation in GP, we modify crossover
and mutation operators so that they produce individuals that maintain
useful intervals. Following crossover or mutation, the
interval of the parent node of the modification site in the offspring
is computed: if this interval is invalid, then a search is
performed to find an operator that will take the child intervals as
input and return a valid output interval. Once a valid operator is
found, the check proceeds up the tree until the root node is
encountered. If an operator cannot be found to produce a valid output
interval, the offspring is flagged as producing an invalid output
interval and assigned a low fitness. This repair mechanism is outlined
in Algorithm~\ref{alg:check-repair}.

\section{Experimental Comparison}\label{sec:benchmarks}

To test the performance of our interval preserving operators, we
selected a range of problems from previous work. Two problems, {\em
  Keijzer-13} and {\em Pagie-1} are synthetic in nature --- {\em
  Keijzer-13} has proven very difficult for standard GP in previous
work, while the {\em Pagie-1} problem has been selected as its generating
function:\begin{equation}
  f\left(x_{1}, x_{2}\right)=\frac{1}{1+x_{1}^{-4}} + \frac{1}{1+x_{2}^{-4}}
\end{equation}
has all its input features in the denominators, and their intervals
include zero. Therefore, this should prove to be a difficult function
for interval arithmetic. The other six functions have been used in
previous work to test genetic programming and other machine learning
methods~\cite{friedman1991multivariate,quinlan1993combining,breiman2001random,Agapitos2014,dick2015improving}. All
parameter settings used in these experiments are the same used earlier
as outlined in Table~\ref{tab:parameters}. For the the {\em
Keijzer-13} problem, we generated a mesh of points as outlined in
previous work, and selected 20 points from this mesh for training for
each run, with the remaining points used for testing. For the {\em
Pagie-1} problem, we again used a mesh over the two variables to
sample 676 points from the problem domain. We then sampled 68 points
(\~10\%) for training and used the rest for testing. For the remaining
problems, we used 10 rounds of 10-fold cross validation to generate
training and testing splits. Each combination of algorithm and problem
was therefore run 100 times. For the {\em Keijzer-13} and {\em
Pagie-1} the known intervals for the input variables was used: for the
other problems, input intervals were estimated by examining the range
of input variables in the training data. The impact of estimating
input intervals will be examined later in this section.

The training performance of the various approaches on the eight
problems is shown in Figure~\ref{fig:bench-train}. As can be seen, for
most of the problems, the revised search operators that consider the
input intervals allow the GP system to evolve at a faster rate than
when using interval arithmetic solely to evaluate individuals. With
the exception of the {\em Concrete} and {\em Friedman-3} problems,
this rate is comparable to the rate seen with protected operators. The
performance of the interval-aware operators on {\em Pagie-1} is
surprising and in contrast to simple interval arithmetic. As expected,
using interval arithmetic solely for evaluation increases the
difficulty, and search performance is restored once interval-aware
operators are adopted.

\begin{figure*}
  \input{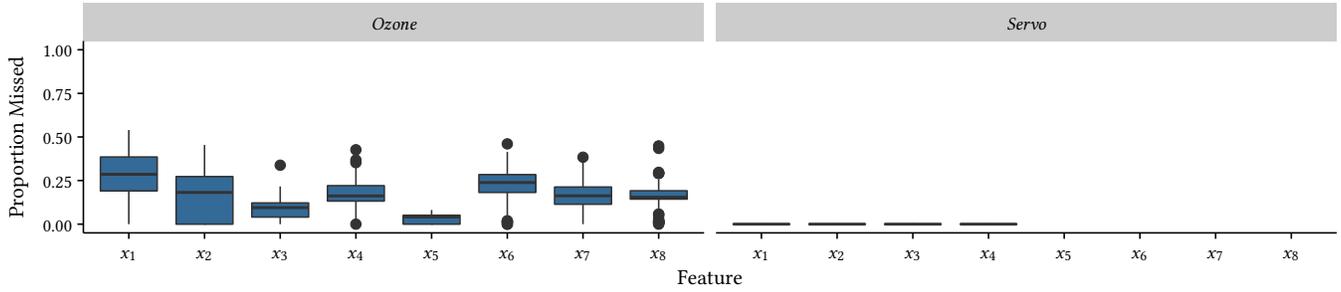}
  \caption{Misalignment between training and testing input intervals
    on the {\em Ozone} and {\em Servo} problems.}
  \label{fig:overlap}
\end{figure*}

\begin{figure*}
  \input{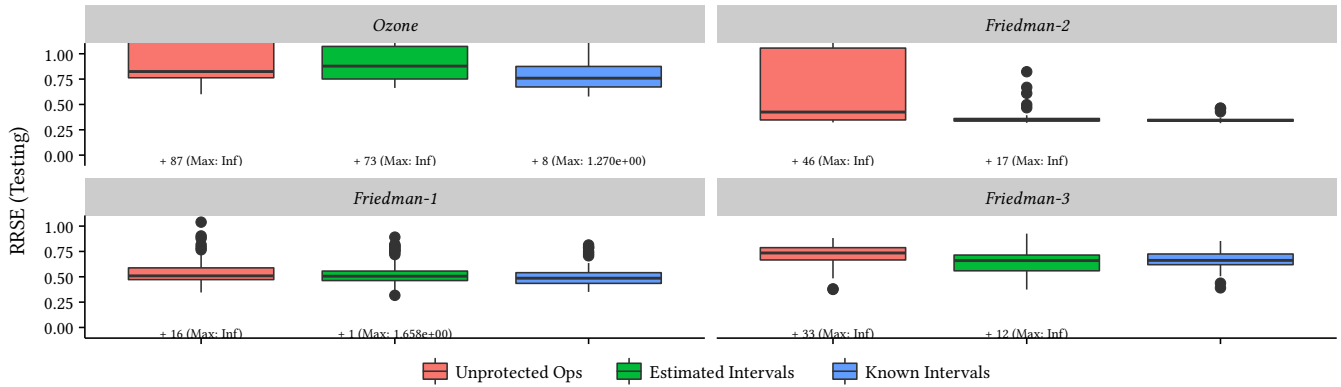}
  \caption{Training performance of interval arithmetic once measured
    (rather than estimated) intervals are provided.}
  \label{fig:fullinterval-final}
\end{figure*}

The testing performance of the examined methods on the eight test
problems is shown in Figure~\ref{fig:bench-test}. The test performance
is rather interesting. In particular, it is surprising that interval
arithmetic, and interval-aware operators, despite having access to the
input intervals, are not able to adequately control generalisation
performance on three problems ({\em Ozone}, {\em
  Friedman-2} and {\em Friedman-3}). For the remaining problems,
interval-aware operators generally present a small but useful
improvement over straight interval arithmetic.

When looking globally, and considering only the median performance of
the algorithms, there is some evidence that there is a difference
between the methods, confirmed by a Friedman test where median
algorithm performance is grouped by method and blocked by problem
($\chi^2=9.1519$, 3 degrees of freedom, p = 0.02734). However, looking
at the boxplots suggests that, with the exception of the {\em Ozone}
problem, methods including interval arithmetic are doing a better job of
controlling destructive overfitting in most cases.

\subsection{Sensitivity of Using Extimated Intervals}\label{sec:sensitivity}
The performance of interval-based methods in the previous experiment
was surprising. This is particularly so in the {\em Ozone} problem,
where no method could adequately control the generalisation
performance. Previous work suggests that interval arithmetic is
guaranteed to provide correct interval information, but this naturally relies
on the supplied intervals being an accurate reflection of what might
be encountered outside of the training process. This suggests that the
estimation process used to identify input intervals in the previous
experiment is inadequate. Evidence towards supporting this is
presented in Figure~\ref{fig:overlap}, which presents the proportion
of the intervals of the input features that was not covered by the
training data for the {\em Ozone} and {\em Servo} problems. These two
problems present the two extremes of conditions encountered in the
previous experiment. As can be seen, the intervals estimated from the
training data in the {\em Ozone} problem are typically poor
representations of the true intervals, whereas the intervals estimated
within the {\em Servo} problem are well-aligned with the intervals
that span testing.

To further examine the influence of the input intervals on algorithm
performance, the four problems that were most problematic in the
previous experiment were re-run with intervals that were either
identified in previous work or measured from the entire data set
rather than just the training
data~\cite{friedman1991multivariate}. The results of this are shown in
Figure~\ref{fig:fullinterval-final}. As can been seen, the performance
of the interval-aware operators is greatly improved from that shown in
Figure~\ref{fig:bench-test}. In all cases, destructive overfitting has
been eliminated --- there are a handful of poor fitting models
remaining for the {\em Ozone} problem, but these are due to poor
search being performed in those runs, rather than through inadequate
handling of intervals.

\section{Conclusion and Future Work}\label{sec:conc}
Symbolic regression has the potential to be a useful method for
machine learing and data science. Traditionally, it has required the
use of protected operators, and this has impacted on the quality of
solutions evolved through GP. Previous work advocated the use of
interval arithmetic to eliminate the need for protected
operators. This paper extended previous work to uncover the dynamics
of interval arithmetic during the evolutionary process. New operators
were developed to greatly reduce the number of invalid
solutions generated during a run, which allows evolution to
proceed at a greater rate then by using interval arithmetic solely in
the evaluation of individuals. Additionally, it was shown that
operators using interval arithmetic are sensitive to the quality of
the estimates of the input intervals --- if intervals can be
identified up front, then interval arithmetic is a reliable and safe
way to conduct symbolic regression search. However, if there is poor
alignment between the intervals used for training models and the
intervals encountered subsequent to training, then interval arithmetic
cannot greatly improve the generalisation performance of symbolic
regression. Therefore, if interval arithmetic is to be used, then
consideration must be made towards adequate identification of input
intervals. In many cases, these intervals should not be difficult to
establish --- many measurements naturally lend themselves to
reasonable identification of valid intervals (e.g., a person's age
cannot be negative, and is not known to exceed 123 years). Therefore,
the practical implications of this limitation of interval arithmetic
are probably not as severe as demonstrated in this paper.

\vspace{-1em}\subsection{Future Work}
The work presented in this paper opens several opportunities for
future research. First, this paper only explored pure interval
arithmetic for use in static analysis and operator execution. However,
straight interval arithmetic is known to possess some issues in
identifying true intervals. For example, if the interval of a variable
$x$ is [0, 1], then the interval of $x - x$ using straight interval
arithmetic is [-1, 1], even though intuitively we know it is [0,
  0]. Additionally, interval arithmetic is unable to identify
correlations between variables, so the resulting intervals that it
identifies are often much wider than would be encountered in
practice. Affine arithmetic has been proposed as an alternative to
interval arithmetic that considers variable correlations, and has been
explored in the context of GP for static analysis~\cite{de2004affine,pennachin2010robust}. While we did
not consider affine arithmetic in this work, it would be a useful
exercise for future work to explore its use in the self-reparing
operators we have developed.

Beyond a rudimentary analysis of the {\em Keijzer-10} problem, We have
not performed a thorough analysis of the structure of the models that
are produced with and without interval arithmetic. It would be
interesting to compare the differences between these --- it is likely
that the search space is altered significantly through the integration
of interval arithmetic, so it should make certain forms of operation
(e.g., division) more difficult to identify. For example, the results
on the {\em Pagie-1} problem using interval-aware operators are
interesting and somewhat counter to what was expected. Interval
arithmetic should find this a very difficult problem to search, as the
input variables appear in the denominator of the problem, and the
intervals associated with these inputs crosses zero (even if the input
data never includes zero itself). However, despite this, the
interval-preserving operators were able to perform at a level
comparable to protected operators on this problem, suggesting that a
suitable surrogate form was being evolved to replace the
division.

Semantic operators are currently an active area of research in
GP. There appears to be alignment between semantic methods and
the interval-preserving operators in this work. Both attempt to
idenfity properties of the subtrees in solutions and use these
properties to guide the search process. More work exploring the
alignment of these two concepts could potentially yield effective
operators that allow GP to be applied to more difficult problems.

Finally, this paper used interval arithmetic for static analysis and
repairing individuals corrupted by search operators. However, there
are other ways that interval analysis could be used within GP that
future work could explore. For example, previous work explored using
the outputs of execution on data to identify areas in models that
could be simplified into constant terms (numerical
simplification)~\cite{kinzett2009numerical}. This allows aspects of trees to be simplified in a
way that the alternative algebraic simplification cannot identify. Use
of interval arithmetic (or, more likely affine arithmetic) within
static analysis could potentially identify areas for simplification
that would normally require the use of both algebraic and numerical
simplication.

\bibliographystyle{ACM-Reference-Format}

\end{document}